\documentclass[preprint,5p,twocolumn]{elsarticle}
\makeatletter
\def\ps@pprintTitle{%
 \let\@oddhead\@empty
 \let\@evenhead\@empty
 \def\@oddfoot{\hfill\thepage\hfill}%
 \let\@evenfoot\@oddfoot}
\makeatother

\usepackage{amsmath}
\usepackage{amsthm}
\usepackage{amsfonts}
\usepackage{mathtools}
\usepackage{amsthm}
\usepackage{amssymb}
\usepackage{nicefrac}
\usepackage{marvosym}
\usepackage{bm}

\usepackage{url}
\usepackage[hidelinks]{hyperref}

\usepackage{tikz}
\usetikzlibrary{arrows.meta,positioning,fit,calc}
\usetikzlibrary{shapes.geometric, arrows, positioning}
\usetikzlibrary {backgrounds}

\usepackage{color}
\definecolor{RWTHBlue}{rgb}{0, 0.32941176470588235, 0.6235294117647059}
\definecolor{RWTHBordeaux}{rgb}{0.6313725490196078, 0.06274509803921569, 0.20784313725490197}
\definecolor{RWTHRed75}{rgb}{0.84705882352941175,0.36078431372549019,0.25490196078431371}
\definecolor{RWTHRed}{rgb}{0.8,0.027450980392156862,0.11764705882352941}
\definecolor{RWTHRed50}{rgb}{0.90196078431372551,0.58823529411764708,0.47450980392156861}
\definecolor{RWTHRed25}{rgb}{0.95294117647058818,0.803921568627451,0.73333333333333328}
\definecolor{RWTHOrange}{rgb}{0.9647058823529412, 0.6588235294117647, 0}
\definecolor{RWTHGreen}{rgb}{0.3411764705882353, 0.6705882352941176, 0.15294117647058825}
\definecolor{RWTHViolett}{rgb}{0.38039215686274508, 0.12941176470588237, 0.34509803921568627}
\definecolor{RWTHTuerkis}{rgb}{0.0, 0.596078431372549, 0.63137254901960782}
\definecolor{RWTHPetrol}{rgb}{0.0, 0.38039215686274508, 0.396078431372549}

\usepackage{xspace}
\usepackage{xifthen}

\newcommand*{\dims}[1][]{
    \ensuremath{
        \ifthenelse{\isempty{#1}}
            {d}
            {d_{#1}}
    }
}

\newcommand*{\R}{\ensuremath{\mathbb{R}}}

\newcommand*{\trans}{\top}
\DeclareMathOperator{\vect}{vec}

\newcommand{\norm}[1]{\left\lVert#1\right\rVert}

\newcommand*{\act}{\ensuremath{u}\xspace}
\newcommand*{\Act}{\ensuremath{U}\xspace}

\newcommand*{\state}{\ensuremath{x}\xspace}
\newcommand*{\States}{\ensuremath{X}\xspace}

\newcommand*{\dist}{\ensuremath{w}\xspace}
\newcommand*{\Dist}{\ensuremath{W}\xspace}
\newcommand*{\distsp}{\ensuremath{\mathcal{W}}\xspace}
\newcommand*{\traj}{\ensuremath{\bm{\tau}}\xspace}

\newcommand*{\step}{\ensuremath{k}\xspace}

\newcommand*{\credible}{\alpha}
\newcommand{\risk}{\epsilon}

\newcommand{\fig}{Fig\/.\/~}

\newcommand{\ie}{i\/.\/e\/.,\/~}
\newcommand{\eg}{e\/.\/g\/.,\/~}
\newcommand{\cf}{cf\/.\/~}

\newtheorem{lemma}{Lemma}
\newtheorem{theorem}{Theorem}

\newtheorem{definition}{Definition}
\newtheorem{assumption}{Assumption}

\newtheorem{remark}{Remark}

\newcommand{\fakepar}[1]{\vspace{1mm}\noindent\textbf{#1.}}

\usepackage[acronym]{glossaries}
\glsdisablehyper
\newacronym[]{lmi}{LMI}{Linear Matrix Inequality}

\journal{Systems \& Control Letters}

\begin{document}

\begin{frontmatter}

\title{Robust Direct Data-Driven Control for Probabilistic Systems}

\author[dsme]{Alexander von Rohr}
\author[dsme]{Dmitrii Likhachev}\author[dsme]{Sebastian Trimpe}

\affiliation[dsme]{organization={Institute for Data Science in Mechanical Engineering, RWTH Aachen University},%
            city={Aachen},
            country={Germany}
}

\begin{abstract}
We propose a data-driven control method for systems with aleatoric uncertainty, for example, robot fleets with variations between agents. Our method leverages shared trajectory data to increase the robustness of the designed controller and thus facilitate transfer to new variations without the need for prior parameter and uncertainty estimations. In contrast to existing work on experience transfer for performance, our approach focuses on robustness and uses data collected from multiple realizations to guarantee generalization to unseen ones. Our method is based on scenario optimization combined with recent formulations for direct data-driven control. We derive lower bounds on the amount of data required to achieve quadratic stability for probabilistic systems with aleatoric uncertainty and demonstrate the benefits of our data-driven method through a numerical example. We find that the learned controllers generalize well to high variations in the dynamics even when based on only a few short open-loop trajectories. Robust experience transfer enables the design of safe and robust controllers that work ``out of the box'' without any additional learning during deployment.
\end{abstract}

\end{frontmatter}

\section{Introduction}\label{sec:intro}

Data-driven control uses data collected from the system for the control design. The predominant assumption for most methods is that the data is collected on a single system and this system is the one being controlled later on. While under laboratory conditions this is often true, in practice the same controller is may be deployed on many realizations of similar systems. For example, in robotic fleets for forest fire fighting \cite{haksar2018distributed} or agriculture \cite{emmi2014new}, and in power electronics for wind farms \cite{markovsky2023data}.
However, the different instantiations of such systems are rarely perfectly homogeneous and each system is subject to variations, for example due to production and assembly procedures or different setups and payloads.
In this case, designing specific controllers for all possible configurations can quickly become impractical.
Instead, it is desirable to design a controller that is robust to this variability and can be used out of the box. 

This is a classical use case of robust control methods.
However, these methods assume that a nominal system model and an uncertainty description are known during control design.
In contrast, data-based control (partly) replaces model knowledge with data and can be designed to be robust to uncertainty after learning on finite and noisy data \cite{berkenkamp2015safe,umlauft2018uncertainty,rohr2021probabilistic,jiao2022backstepping}.
Yet, these method are developed for learning on a single system and are not necessarily robust to variations between multiple systems and have no guarantees for new variations that were not encountered during learning.
We propose a data-based control synthesis that shares data between multiple instantiations and yields controllers that provably generalize to unseen variations.
In summary, our proposed method yields robustness guarantees for data-driven control of probabilistic systems without the need to learn on all instantiations.

\begin{figure}[t]
\centering
    \input{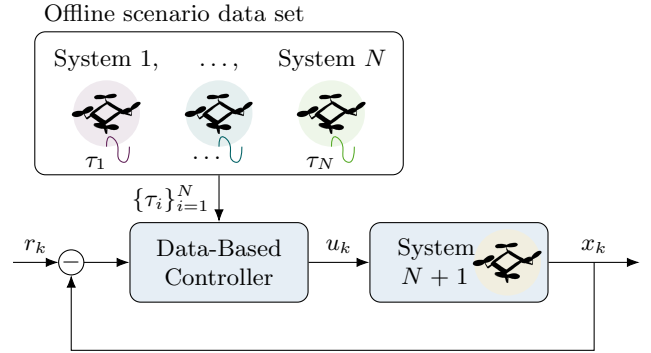}
    \caption{\textbf{Sketch of the proposed method}: A controller is based on a set of collected trajectory data from $N$ systems with variations (here symbolized as quadcopters). The resulting controller is guaranteed to work on any unseen system drawn from the same probability distribution with high probability.}
    \label{fig:fig1}
\end{figure}

We envision our method is especially useful for learning-based control methods that benefit from a preliminary stabilizing control for learning and data collection. The proposed method guarantees generalization and therefore data can be collected once on a subset of systems and be re-used on systems that were not available during the learning phase. Learning on these new systems then can start with an already stabilizing controller.

In this letter, we consider linear dynamics and propose a direct data-driven control\footnote{A `direct' data-driven method forgoes the modeling step, where the dynamics of the open-loop system are identified (\cf \cite{depersis2020formulas}). Instead, the collected data (state-input trajectories) is directly used in the control design.} design for probabilistic systems.
We aim to obtain a single state-feedback controller for all possible variations, including ones that are unseen during the data-collection phase (\cf \fig\ref{fig:fig1}).
Formally, we consider variations as a realization of a random variable influencing the system's dynamics, and we consider robustness as a probabilistic property of the closed-loop system; hence, we aim to design a controller that is stable with high probability with respect to (w.r.t.) the randomness in the variations.
For the probabilistic control design, we employ the scenario approach \cite{campi2009scenario}.
Our proposed combination of the scenario approach with direct data-driven control treats an observed trajectory as an \emph{uncertain scenario}.
We make the following contributions: 
\begin{enumerate}
    \item We introduce the concept of probabilistic data informativity for stabilization as an extension to the informativity framework of \citet{waarde2020data}.
    \item We propose a method to compute a stabilizing controller based on informative data and derive lower bounds on the required number of trajectory samples based on the scenario approach.
    \item We further demonstrate the effectiveness and benefits of the proposed synthesis on a numerical benchmark example.
\end{enumerate}

\subsection{Related Work}\label{sec:related_work}

This section discusses the related literature on cross-system experience transfer, learning-based control designs that consider probabilistic robustness, and recent advances in direct data-driven control.

\fakepar{Experience Transfer for Learning-Based Control} 
In this letter, we present an approach for probabilistic systems where we use trajectories of sampled systems as data. This can be seen as a type of data sharing or experience transfer between the different realizations.
A related body of work on experience transfer uses data collected from variations of source systems to derive a controller on a target system to improve control performance on this target.
In the framework of iterative learning control (ILC) it has been shown that experience transfer can lead to performance increases, but only in limited cases \cite{schoellig2010independent, schoellig2012limited}.
\citet{sorocky2020share} investigate the conditions for an increased performance after experience transfer for the case of linear single-input single-output systems using deep neural networks as transfer functions.
In contrast to these works on transfer learning to increase performance, we use the diversity of the data for the complementary goal of robustness.

Our approach is also related to domain and dynamics randomization developed in the machine learning community to enable deep reinforcement learning (RL) to generalize to real environments from simulated ones (see \citet{muratore2022robot} for a recent survey).
Basically, domain randomization generates artificial environmental or dynamics variations during training of an RL policy and to robustify the trained policy.
Instead of artificial variations of a simulator, we use trajectories collected from different realizations, and we provide controllers that are guaranteed to stabilize variations sampled from the underlying distribution with high probability.

\fakepar{Probabilistic Robust Control}
Generally, considering uncertainty about the underlying dynamics is not new and the topic of robust optimal control \cite{zhou1996robust}.
Robust control methods require an accurate description of the nominal system and its uncertainties to guarantee robustness and performance.
Learning-based control often combines robust control methods with machine learning to get models and their uncertainties from data.
The uncertainty descriptions result from statistical learning theory \cite{koller2018learning, umlauft2018uncertainty,helwa2019provably,fiedler2021learning} or probabilistic modeling \cite{berkenkamp2015safe, umenberger2018learning, rohr2021probabilistic, jiao2022backstepping}.
These methods account for the uncertainty inherent in learning from finite and noisy data, the epistemic uncertainty. 
The goal in our problem formulation is to learn from data from multiple realizations of the probabilistic uncertainty and account for the additional uncertainty from the variations, the aleatoric uncertainty.

Robust control methods are often conservative.
The scenario approach provides a solution by relaxing a given constraint set to a randomly sampled subset and providing probabilistic guarantees \cite{campi2009scenario}. However, designing the distribution over the sampled constraints is still difficult a priori. In the case of a single deterministic system, one can use probabilistic machine learning to infer a distribution over systems to generate scenarios \cite{umenberger2018learning, rohr2021probabilistic}. Direct data-based control treats a trajectory as a single data point without identifying a model, resulting in our scenario formulation that directly incorporates data from the systems. Our proposed approach considers the uncertainty present in a probabilistic system without prior knowledge on distribution or the uncertainty set.

\fakepar{Direct Data-Driven Control} Direct data-driven control is an emerging approach to learning-based control based on \emph{Willems' Fundamental Lemma} \cite{willems2005note}, which characterizes the behavior of a linear system as linear combinations of columns from a data matrix based on trajectory data.
In prior work on direct data-driven control, robustness is defined w.r.t. uncertainties in the data due to unknown disturbances.
One can achieve robustness through regularization \cite{depersis2020formulas,depersis2021low,doerfler2022role,doerfler2023bridging,doerfler2023certainty} or by upper-bounding the process noise \cite{berberich2019robust,berberich2022combining, bisoffi2021trade}.
Herein, we focus on the latter, specifically on a formulation proposed by \citet{waarde2022noisy}.
However, our proposed approach is compatible with both robustification approaches and most direct data-driven control formulations we are aware of.
Our contribution adds to this body of work by considering data collected from different realizations of a probabilistic system and a controller that generalizes well, meaning there is no need to collect data from every possible variation.
While existing formulations are robust w.r.t. \emph{uncertain data}, they do not provide any explicit robustness w.r.t. \emph{uncertain systems}.
The data uncertainty formulations typically provide \emph{less} robust controllers w.r.t. variations as more data is collected. This is desired when learning for a single system because with more data the uncertainty shrinks and less robustness is needed. 
In contrast, our formulation becomes \emph{more} robust with data from different systems.
An approach close to our contribution considers stochastic systems in a distributionally robust framework \cite{coulson2022distributionally}.
The authors present a chance constraint model predictive control formulation, where multiple trajectories are sampled to estimate a distribution over disturbances.
Their resulting controller is then distributionally robust w.r.t. the disturbances.
In contrast, we consider the systems themselves as stochastic realizations, and instead of an MPC design, we propose a probabilistically-robust state-feedback controller.

\section{Problem Formulation}

We consider a probabilistic linear, time-invariant and discrete-time system
\begin{equation}\label{eq:linear_dynamics}
    \state_{\step+1} = A \state_{\step} + B \act_{\step} + \dist_\step,
\end{equation}
where $\state_\step \in \R^{\dims[\state]}$ is the state, $\act_\step \in \R^{\dims[\act]}$ is the input and $\dist_\step  \in \R^{\dims[\state]}$ is the process noise of the system.
In our problem formulation the system matrices $A$ and $B$ are \emph{random variables} with an unknown distribution.
We make the standard assumptions for direct data-driven control: noise-free measurements of the state $\state_\step$, and i.i.d. process noise $\dist_k$ \cite{berberich2019robust,depersis2020formulas,waarde2020data}.
A sample from the probabilistic system is defined by the parameter tuple $\theta_i = (A_i, \; B_i)$.
\begin{definition}[Probabilistic linear system]\label{def:fleet}
    A probabilistic system is a random variable $\Theta: \Omega \to \hat\Theta$ with domain $\hat \Theta \subseteq \R^{\dims[\state] \times (\dims[\state] + \dims[\act])}$ on a probability space $(\Omega, \mathcal{F}, \mathbb{P})$.
    All variations $\theta_i \in \hat\Theta$ are i.i.d. samples from $\Theta$.
\end{definition}
This definition is a formalization of system variations as realizations of a random variable. It allows us to define the control synthesis problem in a probabilistic framework. Note here, that the random variable is not a function of the time step $k$, \ie a realization of the probabilistic system is fixed in the closed-loop.
We aim to solve such problems without any knowledge about the parameters of the variations $\theta_i$, the random variable $\Theta$ and its distribution, nor its domain $\hat\Theta$.
Instead, we have incomplete access to the behavior of the probabilistic system through data in the form of a finite number of $N$ state-input trajectories.
For the control synthesis problem to be feasible, we need to assume the distribution of the random variable is not `too wide' so that there exists a single state-feedback controller $K$ that can stabilize the probabilistic system with high probability.
\begin{definition}[$\credible$-probabilistic robust controller]\label{def:apr}
    A state feedback controller $K$ is $\credible$-probabilistic robust w.r.t. $\Theta$ if there exists a $P=P^\trans \succ 0$ such that
    \begin{equation}
        \mathbb{P}_{\Theta}\left[ P - (A+BK)P(A+BK)^\trans \succ 0 \right] \geq 1 - \credible.
    \end{equation}
\end{definition}%
\begin{assumption}\label{ass:stabilizable}
    We assume the probabilistic system is $\credible$-probabilistically stabilizable, \ie there exists a state feedback controller $K$ that is $\credible$-probabilistic robust w.r.t. $\Theta$.  
\end{assumption}

The data-driven control design problem is now as follows: After seeing trajectories from $N$ different variations of length $M+1$, denoted as 
\begin{equation}
    \traj_i=((\state_{0,i},\state_{0,i}),\dots,(\state_{M,i},\act_{M,i})),
\end{equation}
from systems that are sampled i.i.d. from $\Theta$, we want to find an $\credible$-probabilistic robust controller. 

\section{Preliminaries}

We define the following matrices for the trajectories $\traj_i$
\begin{align}
    \States_i &= \begin{bmatrix} \state_{0,i} & \state_{1,i} & \dots & \state_{M-1,i} \end{bmatrix}, \\
    \States^+_i &= \begin{bmatrix} \state_{1,i} & \state_{2,i} & \dots & \state_{M,i} \end{bmatrix}, \\
    \Act_i &= \begin{bmatrix} \act_{0,i} & \act_{1,i} & \dots & \act_{M-1,i} \end{bmatrix}, \\
    \Dist_i &= \begin{bmatrix} \dist_{0,i} & \dist_{1,i} & \dots & \dist_{M-1,i} \end{bmatrix}
\end{align}

\subsection{Data Informativity in Direct Data-Driven Control}

A fundamental question when designing controllers from data is: What data is needed in order to guarantee a desired outcome?
To answer this question, the informativity framework \cite{waarde2020data} formalizes assumptions on the data and model class to analyze systems and design controllers from data.
Given a model class $\Sigma$, \eg all linear systems of a certain state-input dimension
\begin{equation}
    \Sigma=\{\theta = (A,B)\mid A \in \R^{\dims[\state] \times \dims[\state]},\: B \in \R^{\dims[\state] \times \dims[\act]}\},
\end{equation}
and a trajectory $\traj$, we define a set $\Sigma_\tau \subseteq \Sigma$ as the set of all systems that are consistent with the data, \ie all systems that could have produced the observed trajectory.
Since we are interested in stability, we define the set $\Sigma_K$ as the set of all systems stabilized by the static feedback gain $K$.

A data set is informative about stabilization if there exists a controller $K$ such that $\Sigma_\tau \subseteq \Sigma_K$: the set of systems stabilized by $K$ contain the whole uncertainty set $\Sigma_\tau$.
Clearly, if there are no assumptions on the disturbances $W$, then data cannot be informative and $\Sigma_\tau = \Sigma$.
Therefore, the need arises to formulate an assumption, which often is taken as a bound on the disturbances.
We follow \citet{waarde2022noisy} and formulate such bound in the form of a set membership for the disturbance trajectory.
\begin{assumption}[{\cite[Assumption~1]{waarde2022noisy}}]\label{ass:noise}
    The matrices $\Dist_i$ are an element of $\distsp$, where 
    \begin{equation}\label{eq:W}
        \distsp = \left\{ \Dist \in \R^{\dims[\state] \times M} \, \middle| \, \begin{bmatrix}
            I \\ \hat \Dist^\trans
        \end{bmatrix}^\trans
        \begin{bmatrix}
            \Phi_{11} & \Phi_{12} \\
            \Phi_{12}^\trans & \Phi_{22}
        \end{bmatrix}
        \begin{bmatrix}
            I \\ \hat \Dist^\trans 
        \end{bmatrix}
        \succeq 0
        \right\}
    \end{equation}
for some known $\Phi_{11} = \Phi_{11}^\trans$, $\Phi_{12}$, and $\Phi_{22} = \Phi_{22}^\trans \prec 0$.
\end{assumption}
If, for example, each noise realization $\dist_k$ is bounded by some known constant $\norm{\dist_k}^2_2 \leq \bar \dist$, then a valid noise model is \eqref{eq:W} with $\Phi_{11} = M \bar \dist I$, $\Phi_{12}$, and $\Phi_{22} = -I$ \cite{waarde2022noisy}.

Equipped with this assumption the uncertainty set containing all systems from the model class $\Sigma$ that are consistent with the data is (\cf\cite[Lemma~4]{waarde2022noisy})
\begin{equation}
    \Sigma_{\tau_i} = \left\{ \theta_i = (A_i, \; B_i) \; \middle| \; \States_i^+ = A_i \States_i + B_i \Act_i + \hat \Dist_i, \hat \Dist_i \in \distsp \right\}.
\end{equation}
By Assumption~\ref{ass:noise}, we have for the true noise realization that $\Dist \in \distsp$. Therefore, the system is in the uncertainty set, \ie $\theta_i \in \Sigma_{\tau_i}$.
This assumption is sufficient to design robust state feedback controllers with noisy data as we will show in the next section.

\subsection{Robust State Feedback}

After the definition of the uncertainty set $\Sigma_{\tau_i}$, we present here the \gls{lmi}-based controller synthesis from \citet{waarde2022noisy}.
First, we state a condition on the data, and second, we use this data to parameterize a state-feedback controller $K$ that stabilizes all systems in the uncertainty set $\Sigma_{\tau_i}$ for deterministic system.

\begin{definition}[Generalized Slater's condition]
    The generalized Slater condition is fulfilled if there exists some matrix $Z \in \R^{(\dims[\state] + \dims[\state]) \times \dims[\state]}$ such that
    \begin{equation}\label{eq:gen_slater}
        \begin{bmatrix}
            I \\ Z
        \end{bmatrix}^\trans
        V
        \begin{bmatrix}
            I \\ Z
        \end{bmatrix}
        \succeq 0, \text{with}
    \end{equation}
    \begin{equation}\label{eq:N}
        V = 
        \begin{bmatrix}
            I & \States^+_i \\
            0 & -\States_i \\
            0 & -\Act_i
        \end{bmatrix}
        \begin{bmatrix}
            \Phi_{11} & \Phi_{12} \\
            \Phi_{12}^\trans & \Phi_{22}
        \end{bmatrix}
        \begin{bmatrix}
            I & \States^+_i \\
            0 & -\States_i \\
            0 & -\Act_i
        \end{bmatrix}^\trans
    \end{equation}
\end{definition}
\begin{remark}
   The condition \eqref{eq:gen_slater} is used a prerequisite that the recorded trajectory data can be used in to check the informativity condition and can be used in the \gls{lmi} for controller synthesis. It can easily be checked and in our empirical evaluation we found that under some mild excitation it is almost always fulfilled. In practice, if a trajectory does not fulfill the condition, it can be discarded and re-recorded. 
\end{remark}

\begin{lemma}[{\cite[Th.~14]{waarde2022noisy}}]\label{lemma:robust_stable}
    Assume the generalized Slater condition \eqref{eq:gen_slater} holds. Then the data $\tau_i$ is informative for quadratic stabilization if and only if there exists $P \in \R^{\dims[\state] \times \dims[\state]}$ with $P = P^\trans \succ 0$, $L \in \R^{\dims[\act] \times \dims[\state]}$ and scalars $a \geq 0$, $b > 0$  satisfying 
    \begin{equation}\label{eq:lmistab}
    \begin{split}
        &\begin{bmatrix}
            P-b I & 0 & 0 & 0 \\
            0 & -P & -L^\trans & 0 \\
            0 & -L & 0 & L \\
            0 & 0 & L^\trans & P
        \end{bmatrix} \\ 
            - a &\begin{bmatrix}
            I & \States^+_i \\ 0 & -\States_i \\ 0 & -\Act_i \\ 0 & 0
            \end{bmatrix}
            \begin{bmatrix}
            \Phi_{11} & \Phi_{12} \\
            \Phi_{12}^\trans & \Phi_{22}
            \end{bmatrix}
            \begin{bmatrix}
            I & \States^+_i \\ 0 & -\States_i \\ 0 & -\Act_i \\ 0 & 0
            \end{bmatrix}^\trans \succeq 0.
    \end{split}
    \end{equation}
    If $P$ and $L$ satisfy \eqref{eq:lmistab}, then $K := L P^{-1}$ is a stabilizing feedback controller for all $\theta \in \Sigma_{\tau_i}$.
\end{lemma}
Lemma~\ref{lemma:robust_stable} shows that a solution to the LMI \eqref{eq:lmistab} yields a controller that is guaranteed to stabilize the closed-loop system from which the trajectory is collected. 
Therefore, the controller synthesis problem can be solved by finding a solution to the above \gls{lmi}.

\section{Stabilization for Finite Sets of Systems}\label{sec:lmi}

\begin{figure}[t]
    \centering
    \vspace{3mm}
    \scalebox{1}{\input{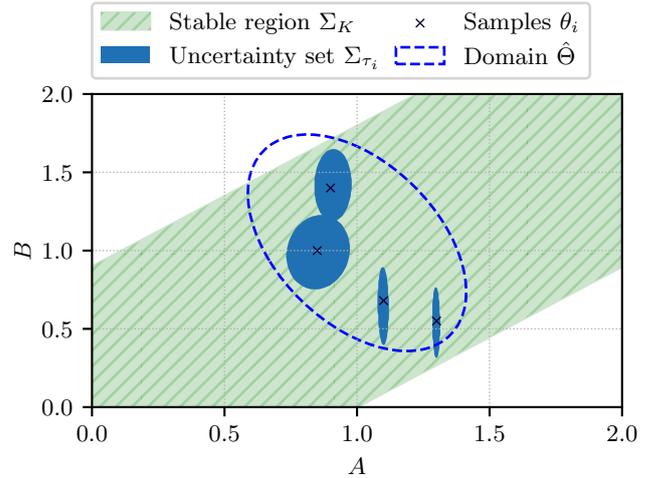}}
    \caption[Illustrative example of scenario optimization for a probabilistic one-dimensional systems]{\textbf{Illustrative example of scenario optimization for a one-dimensional probabilistic system}: Resulting controller robustly stabilizes the union of the sets $\Sigma_{\tau_{1}}\dots \Sigma_{\tau_{4}}$. The more samples are available, the more likely it is that the controller stabilizes a sample from $\Theta$ (\cf Theorem~\ref{th:scenario_rddc}). The fraction of $\Theta$ that remains unstable corresponds to $\credible$.}
    \label{fig:1d-scenario}
\end{figure}

We now extend Lemma~\ref{lemma:robust_stable} to the case where the controller stabilizes not only one but a finite set of systems $\{\theta_i\}^N_{i=1}$.
For that purpose, we will use a single trajectory collected from each system.
Again, we do not explicitly identify the systems, but we derive an \gls{lmi} formulation that depends solely on the collected trajectories.
We start by first defining informativity for quadratic  stabilization of a finite set of systems.

\begin{definition}[Informativity for quadratic  stabilization of finite sets of systems]
    Assume all trajectories in the set $\{\traj_i\}^N_{i=1}$ satisfy Assumption~\ref{ass:noise}.
    We say a set of trajectories $\{\tau_i\}^N_{i=1}$ are informative for quadratic stabilization of the set of systems $\{\theta_i\}^N_{i=1}$ if there exists a state feedback gain $K$ and a matrix $P=P^\trans \succ 0$ such that 
    \begin{equation}\label{eq:quad_stab}
        P - (A_i + B_i K) P (A_i + B_i K)^\trans \succ 0 
    \end{equation}
    for all $\theta_i = (A_i, \; B_i) \in \bigcup_{i=1}^N\Sigma_{\tau_i}$.
\end{definition}
Next we show how we can use the \gls{lmi} \eqref{eq:lmistab} to find such a controller.
\begin{theorem}\label{th:robust_stable_multi}
    Assume that the generalized Slater condition \eqref{eq:gen_slater} is fulfilled for all $\{\tau_i\}^N_{i=1}$.
    If there exists $P \in \R^{\dims[\state] \times \dims[\state]}$ with $P = P^\trans \succ 0$, $L \in \R^{\dims[\act] \times \dims[\state]}$ and scalars $a \geq 0$ and $b > 0$ such that \eqref{eq:lmistab} holds for all $i=1,\ldots,N$, then the data $\{\tau_i\}_{i=1}^N$ is informative for quadratic stability of all systems $\theta_i \in \bigcup_{i=1}^{N} \Sigma_{\tau_i}$.
    Furthermore, $K := L P^{-1}$ is a stabilizing controller for all $\theta_i \in \bigcup_{i=1}^{N} \Sigma_{\tau_i}$.
\end{theorem}
\begin{proof}
    If a solution to the \gls{lmi} in \eqref{eq:lmistab} is satisfied for a single trajectory, the system that generated that data is stable with feedback controller $K$ (Lemma~\ref{lemma:robust_stable}). 
    Consider $\mathcal{LP}_i$ the set of all $L_i$ and $P_i$ that satisfy \eqref{eq:lmistab} for a single $\tau_i$. Then the solution to the \gls{lmi} described by \eqref{eq:lmistab} for all $i=1,\ldots,N$ is the intersection of all $\mathcal{LP}_i$.
    Therefore, the $K = L P^{-1}$ must stabilize all sets $\Sigma_{\tau_i}$.
\end{proof}
The \gls{lmi} in Theorem~\ref{th:robust_stable_multi} can be solved using standard solvers.
\fig\ref{fig:1d-scenario} depicts an example of a controller that simultaneously stabilize multiple system by stabilizing the union of the uncertain regions $\bigcup_{i=1}^{N} \Sigma_{\tau_i}$.
Theorem~\ref{th:robust_stable_multi} considers informativity for quadratic stabilization, but readily extends to synthesis problems with additional performance criteria such as the $\mathcal{H}_\infty$ and $\mathcal{H}_2$ formulations in \cite{waarde2022noisy}. 

For this section we have only considered stability for the systems that generated trajectory data. If we want to stabilize a continuous set $\hat \Theta$, this would require solving a semi-infinite optimization problem and collecting an infinite amount of data.
Next, we will show how to find an $\credible$-probabilistically robust controller for the whole distribution with finite data.

\section{Stabilization of Probabilistic Systems}\label{sec:rddc_scenario}

In Sect.~\ref{sec:lmi}, we introduced a sufficient condition for a controller that provably stabilizes a finite set of systems for which there is available data.
To make this result usable for probabilistic systems with possibly infinite number of variations, we show that the solution to the finite program generalizes to the whole distribution using scenario optimization \cite{campi2009scenario}.
First, we define an informativity notion for $\credible$-probabilistic stabilization (\cf Def.~\ref{def:apr}).

\begin{definition}[Informativity for $\credible$-probabilistic quadratic stabilization]
    Assume all $\{\tau_i\}_{i=1}^N$ satisfy Assumption~\ref{ass:noise}.
    Assume further all trajectories are created from i.i.d. samples of the probabilistic system $\Theta  = [A, \; B]$ (Def.~\ref{def:fleet}).
    We say a set of trajectories $\{\tau_i\}^N_{i=1}$ is informative for $\credible$-probabilistic quadratic stabilization of the probabilistic system $\Theta$ if there exists a state feedback gain $K$ and a matrix $P=P^\trans \succ 0$ such that 
    \begin{equation}
        \mathbb{P}_{\Theta}\left[P - (A + BK) P (A + BK)^\trans \succ 0 \right] \geq 1 - \credible.
    \end{equation}
\end{definition}

Next, we show how to determine a lower bound on the number of trajectories required to design an $\credible$-robust controller with high probability. 
\begin{theorem}\label{th:scenario_rddc}
    Assume that the generalized Slater condition \eqref{eq:gen_slater} is fulfilled for all $\{\tau_i\}^N_{i=1}$.
    Select a confidence parameter $\risk \in (0,1)$ and a violation parameter $\credible \in (0,1)$.
    If $N \geq \frac{2}{\credible} (\ln{\frac{1}{\risk}} + n)$ with $n=\dims[\state]^2 + \dims[\state] \cdot \dims[\act] + 2$ and there exist $P = P^\trans \succ 0$, $L \in \R^{\dims[\act] \times \dims[\state]}$ and scalars $a \geq 0$ and $b > 0$ such that \eqref{eq:lmistab} holds for all $i=1,\ldots,N$, then the data $\{\tau_i\}_{i=1}^N$ is informative for $\credible$-probabilistic quadratic stabilization with probability $1-\risk$, \ie
    \begin{equation}
    \begin{split}
        \mathbb{P}_{\tau} \left[\mathbb{P}_{\Theta}\left[ P - (A + BK) P (A + BK)^\trans \succ 0  \right]  \geq 1 - \credible\right] \\  \geq 1 -\risk.
    \end{split}
    \end{equation}
\end{theorem}
\begin{proof}
    The \gls{lmi} \eqref{eq:lmistab} for all $i=1,\ldots,N$ is a convex scenario program with $n=\dims[\state]^2 + \dims[\state] \cdot \dims[\act] + 2$ decision variables. The result follows from \cite[Theorem~1]{campi2009scenario}.
\end{proof}

Note the nested probabilities in Theorem~\ref{th:scenario_rddc}.
The outer probability depends on the sampled data and quantifies the possibility $\risk$ that the samples are not representative of the underlying distribution in which case the controller might fail to generalize. 
However, since the required number of trajectories $N$ only scales logarithmically in $\risk$ this probability can be chosen relatively low without increasing the required samples too much.
Again, this result can be readily extended to the $\mathcal{H}_\infty$ and $\mathcal{H}_2$ setting in \cite{waarde2022noisy}, in which case we can not only guarantee stability but also a minimum performance level for all systems in the fleet with high probability.

\section{Numerical Example}

\begin{figure}[t]
    \centering
    \input{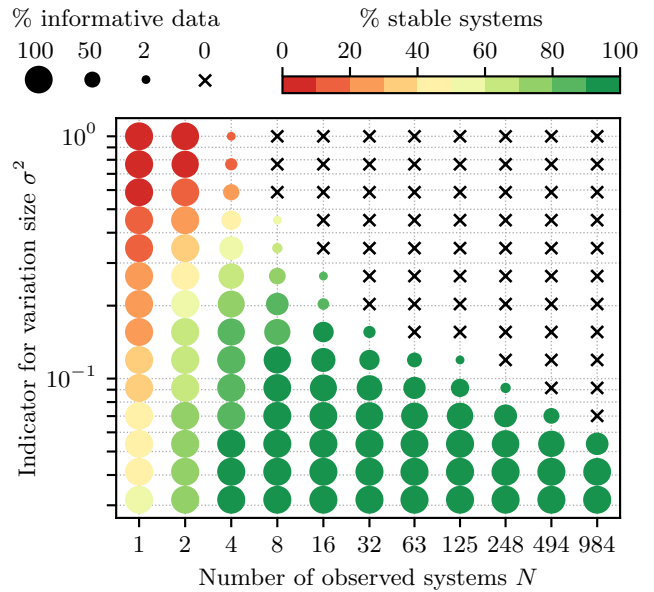}
    \caption[Synthetic example with different number of observed systems and increasing variance of the probabilistic system]{\textbf{Synthetic example with different number of observed systems and increasing variance of the probabilistic system}:
    The probability that the controller stabilizes a sample from the probabilistic system increases as more samples become available.
    For high variance the uncertainty set becomes too large and the synthesis returns no controller ($\times$).
    The frequency of stability is averaged over $50$ different data sets and the size of the dots indicates the average frequency at which the data is informative.
    }
    \label{fig:heatmap1}
\end{figure}

In this section, we apply the results of the scenario optimization for probabilistically robust direct data-driven control on a linear system benchmark problem. 
The theorems developed above guarantee probabilistic stability only if a solution to the \gls{lmi} is found. 
A natural question to ask is: do such \glspl{lmi} have a feasible solution? In this section we give an example for a probabilistic system where they have and show that the proposed method compares favorably to prior work based in probabilistic system identification \cite{berkenkamp2015safe,rohr2021probabilistic}.
Further, we investigate the generalization of our method beyond the lower bound given in Theorem~\ref{th:scenario_rddc} and the influence of the trajectory length on a specific example.
In summary, we find that on the chosen benchmark problem we can deal with relatively high levels of uncertainty and that our method generalizes well even on small data sets.
\footnote{Source code to reproduce results is available at \url{https://github.com/Data-Science-in-Mechanical-Engineering/rddc}.}

\subsection{Probabilistic Linear System}

The synthetic benchmark problem investigated here is adapted from our previous work \citet{rohr2021probabilistic} and is based on a popular example system in the data-based control literature first proposed by \citet{dean2020sample}.
The fleet distribution is
\begin{equation}
    \Theta \sim \mathcal{TN}(\mu, \Sigma_\Theta),
\end{equation}
where $\mathcal{TN}$ is the truncated normal distribution with mean $\mu$ and variance $\Sigma$.
We truncate such that the mean is centered and a sample from the non-truncated normal is inside the interval with a probability of $95\%$.
The mean is chosen as an unstable graph Laplacian system with $\mu  = \vect(\left[\bar A \, \bar B\right])$ with
\begin{equation}\label{eq:dean_dynamics_2}
\bar A =
\begin{bmatrix}
    1.01 & 0.01 & 0\\
    0.01 & 1.01 & 0.01\\
    0 & 0.01 & 1.01
\end{bmatrix} \;
\bar B = I_3.
\end{equation}
For the variance over systems, we choose $\Sigma_\Theta = \sigma^2 (0.5 \cdot I + 0.5 \cdot 1)$, which has a single parameter $\sigma^2$ to control size of the domain $\hat \Theta$ and also allows for correlations between parameters.
The bound of Theorem~\ref{th:scenario_rddc} imposes $N \geq 984$ for $\credible = 0.05$ and $\risk=0.01$.
In our experiments, we first sample $N$ systems from the distribution and use each system to generate a trajectory of length $M+1$ with random initial conditions.
To verify that the controller generalizes to the whole distribution we sample $1000$ new systems from $\Theta$ and test their stability. The percentage of unstable systems is used as an estimate for $\credible$.
We repeat the experiments for all combinations $50$ times and average the probability of closed-loop stability.

\subsection{Generalization and Feasibility of the Synthesis}

In the first experiment we validate the lower bound derived in Theorem~\ref{th:scenario_rddc}.
The trajectory length is set to $M=500$.
The results are shown in \fig\ref{fig:heatmap1}.
As the number of observed systems is increased, the proposed method either yields a controller stabilizing the overwhelming majority of the fleet until the aleatoric uncertainty is too large and no $\credible$-probabilistic robust control can be found.
As expected, for the theoretical lower bound we achieve the desired result that the synthesis returns an $\credible$-probabilistic robust controller.
While using the theoretical lower bound of $N=984$ the synthesis is feasible for distributions uncertainties up to $\sigma^2 \leq 0.05$ and yields no controller for higher uncertainties.
The methods based on probabilistic system identification were able to find controllers for uncertainties up to $\sigma^2 \leq  0.001$ \cite{rohr2021probabilistic} and $\sigma^2 \leq 10^{-4}$ \cite{berkenkamp2015safe} (see the empirical results of \citet{rohr2021probabilistic}).

At least for the benchmark problem the proposed method can deal with higher uncertainties, despite working with uncertain scenarios.  
Furthermore, the controller generalizes even when only $32$ trajectory samples are available. 
In \fig\ref{fig:heatmap2} we can also observe that the relaxed problem with $N=32$ enables synthesis for probabilistic systems with even higher variance and yields controllers for  $\sigma^2 \leq 0.1$, albeit without the guarantees of Theorem~\ref{th:scenario_rddc}.   
These findings are consistent with \citet{umenberger2018learning} that reported learning stabilizing controllers after a few rollouts using probabilistic system identification.

\subsection{Effect of the Trajectory Length}

\begin{figure}[t]
    \vspace{2mm}
    \centering
    \input{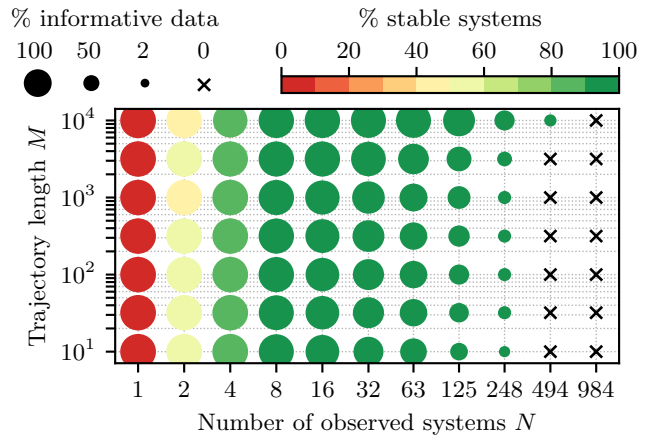}
    \caption[Synthetic example with different number of observed systems and varying trajectory length]{\textbf{Synthetic example with different number of observed systems and varying trajectory length} ($\sigma^2=0.1$):
    The robustness is largely unaffected by the trajectory length but longer trajectories increase data informativity and the method is able to find controllers more often.
    }
    \label{fig:heatmap2}
\end{figure}
\begin{figure*}[ht]
    \centering
    \input{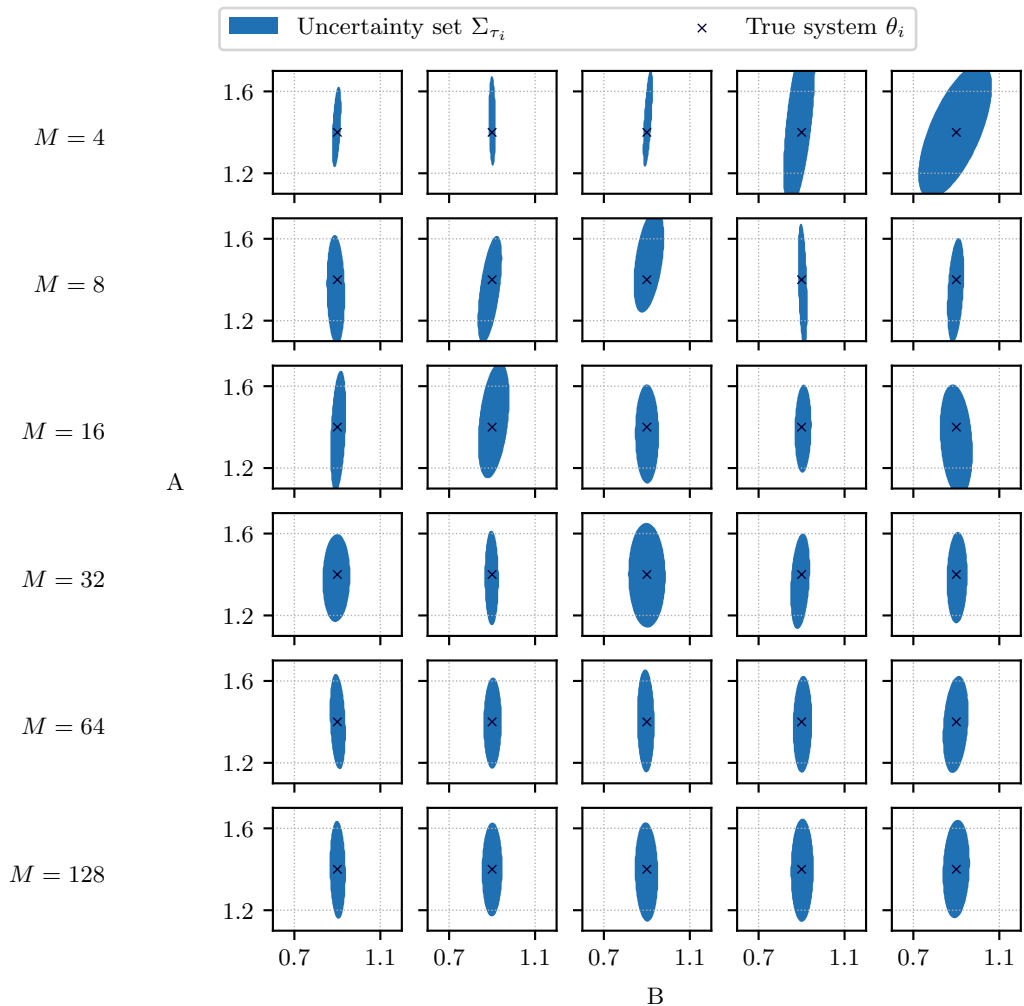}
    \caption[Uncertainty sets for varying trajectory lengths]{\textbf{Uncertainty sets for varying trajectory lengths}: The columns represent rollouts with different noise realizations. For short trajectories, the form, the position, and the size of the uncertainty set vary significantly across different rollouts. Longer trajectories provide more consistent uncertainty sets.}
    \label{fig:sigma_var_T_seed}
\end{figure*}

Our theoretical results we focus on the number of trajectories required to achieve a stabilization controller with high probability. However, the theory does not reveal the effect of the trajectory length. As long as the trajectory satisfies the generalized Slater condition, the result holds. However, since the available data influences the uncertainty sets $\Sigma_{\traj_i}$ our second experiment explores the influence of the trajectory length on the robustness and feasibility of the synthesis problem.
For these experiment we use a fixed distribution $\Theta$ with $\sigma^2=0.1$.
\fig\ref{fig:heatmap2} shows that the trajectory length has no effect on the stability. It does however influence the feasibility of the resulting \gls{lmi}. In our experiment longer trajectories make the problem feasible more often. However, the effect is relatively small and even very short trajectories can be sufficient to achieve stability with high probability.

To investigate the effect of the trajectory length further we show the uncertainty sets $\Sigma_{\traj_i}$ for different values of $M$ in \fig\ref{fig:sigma_var_T_seed}.
For this, we choose the system $A = 0.9, B = 1.4$ and use it to generate trajectories of varying lengths. The process noise bound is $\bar\dist \leq 0.015$. For each trajectory length, we produce $5$ different trajectories with different noise realizations. Then, we calculate the data-consistent set $\Sigma_{\tau_i}$ based on the trajectory and noise assumption.

This figure illustrates the following relationship: Longer trajectories do not decrease the size of the uncertainty sets, but they reduce their variance. The fact that more data does not reduce the uncertainty might first seem counter-intuitive. Usually, when determining an uncertain quantity, performing more measurements reduces the uncertainty, but only down to the measurement tool's tolerance. For determining the data-consistent set, the process noise plays the role of ``measurement tolerance''. When we reach this tolerance, each new sample will not only provide new information but also obfuscate it with the newly introduced noise.

\section{Conclusion}

In this letter, we propose a new approach for direct-data driven control for probabilistic linear system based on the novel concept of probabilistic informativity. 
Our method utilizes a convex scenario program to design controllers that provably generalize to the whole distribution based on trajectory data from realizations of the probabilistic system.
To this extend we provide lower bounds for the necessary numbers of trajectories and demonstrate empirically that the resulting controller synthesis remains feasible even when the variance of the probabilistic system is large.

Further, the method can be used even when the trajectory length is rather short.
This makes the method suitable to devise preliminary stabilizing controllers for learning-based control even for unstable systems where sampling long open-loop trajectories is difficult.
An avenue for future work is developing data-based methods for unstable systems in an iterative fashion, much like reinforcement learning algorithms, with additional performance criterion where the controller is sequentially improved with more samples.



\bibliographystyle{elsarticle-num-names} 
\bibliography{references}

\end{document}